\documentclass[conference]{IEEEtran}
\IEEEoverridecommandlockouts
\usepackage[utf8]{inputenc}
\usepackage{cite}
\usepackage{amsmath,amssymb,amsfonts}
\usepackage{algorithmic}
\usepackage{graphicx}
\usepackage{textcomp}
\usepackage[table]{xcolor}
\usepackage{booktabs}  
\usepackage[section]{placeins} % keep floats from drifting past section boundaries
\usepackage{tikz}
\usepackage{multirow}
\usepackage{makecell}
\usepackage{siunitx}
\usepackage{tabularx}
\usepackage{array}
\usepackage{caption}
\usepackage{diagbox}
\usepackage{threeparttable}
\usepackage[hidelinks]{hyperref}

\def\BibTeX{{\rm B\kern-.05em{\sc i\kern-.025em b}\kern-.08em
    T\kern-.1667em\lower.7ex\hbox{E}\kern-.125emX}}

\definecolor{colW2V2B}{HTML}{ef9a9a}
\definecolor{colW2V2BFT}{HTML}{b71c1c}
\definecolor{colHuBB}{HTML}{90caf9}
\definecolor{colHuBL}{HTML}{1976d2}
\definecolor{colHuBLFT}{HTML}{0d47a1}
\definecolor{colWavLMB}{HTML}{a5d6a7}
\definecolor{colWavLML}{HTML}{2e7d32}
\definecolor{colXLSR}{HTML}{ce93d8}
\definecolor{colMMS}{HTML}{6a1b9a}
\definecolor{maxred}{RGB}{240,130,130}
\definecolor{grpred}{RGB}{255,205,205}
\definecolor{mwu}{HTML}{c62828}
\definecolor{perm}{HTML}{190e7a}
\definecolor{egemaps}{HTML}{FF9800}

\newcommand{\rmaxtext}[1]{\colorbox{maxred}{\textbf{#1}}}
\newcommand{\gmaxtext}[1]{\colorbox{grpred}{\textbf{#1}}}
\newcommand{\rmax}[1]{\cellcolor{maxred}\textbf{#1}}
\newcommand{\gmax}[1]{\cellcolor{grpred}\textbf{#1}} 
\begin{document}

\title{Motor, Cognitive, or Corpus? What Survives Cross-Lingual Transfer in Speech-Based Parkinson's Disease Detection \\
}

\author{
  \IEEEauthorblockN{
   Serli Kopar\textsuperscript{1,2},
    Sam Gijsen\textsuperscript{1,2,3},
    Abner Hernandez\textsuperscript{4},
    Paula Andrea Perez-Toro\textsuperscript{4},
    Kerstin Ritter\textsuperscript{1,2,3}
  }%
  \IEEEauthorblockA{
    \textsuperscript{1}Hertie Institute for AI in Brain Health, University of Tübingen, Tübingen, Germany \\
    \textsuperscript{2}Tübingen AI Center, University of Tübingen, Tübingen, Germany \\
    \textsuperscript{3}Charit\'e--Universit\"atsmedizin, Department of Psychiatry and Psychotherapy, Berlin, Germany
    \\  \textsuperscript{4}Pattern Recognition Lab, Friedrich-Alexander-Universität Erlangen-Nürnberg, Erlangen, Germany
  }
}
\maketitle

\begin{abstract}
Self-supervised learning (SSL) speech representations achieve strong performance for Parkinson's disease (PD) detection within individual corpora. However, it remains unclear whether these models capture disease-related characteristics or exploit dataset-specific confounds, particularly since most SSL backbones are pretrained exclusively on healthy speech. To investigate this question, we perform a layer-wise analysis of nine SSL speech backbones using a low-capacity logistic regression probe across three languages. We structure the evaluation as multiple scenarios that progressively introduce distribution shifts in participant identity, recording conditions, language, and pathology. Our results reveal two key findings. First, layer selection is highly corpus-dependent: the optimal representation layer is determined primarily by the source dataset rather than by the SSL architecture itself. Second, the transferred discriminative signal lacks pathological specificity: classifiers trained to detect PD assign similarly high probabilities to both PD and dementia speech in the target corpus. These results highlight critical limitations that must be addressed before speech-based pathology recognition models can be reliably deployed in clinical settings.

\end{abstract}


%--------------------------------------------------
\section{\textbf{Introduction}}

Speech and language are increasingly used to characterise motor and cognitive changes in Parkinson's disease (PD). Alterations in phonation, articulation, fluency, and lexical access can reflect the combined impact of motor impairment and non-motor symptoms, including cognitive decline \cite{Bloem2021Parkinson}. Clinically, motor severity is assessed using instruments such as the MDS-UPDRS \cite{Goetz2008MDSUPDRS} and Hoehn \& Yahr scale \cite{HoehnYahr1967}, based on speech tasks including sustained vowel phonation (VOWEL), diadochokinetic repetition (DDK), and passage reading (READ). Cognitive impairment is evaluated using neuropsychological batteries such as CERAD-NB+ \cite{morris1989cerad} or screening tools such as the MMSE \cite{Folstein1975}, together with language tasks including Verbal Fluency and the Boston Naming Test.

Automated speech analysis has emerged as a promising complement to these assessments \cite{KOVAC2024105667}. Recent work increasingly relies on self-supervised learning (SSL) speech representations. Most studies either fine-tune the entire SSL model for PD classification \cite{bioengineering12070728} or freeze the pretrained encoder and train a lightweight classifier on the extracted embeddings \cite{Klempir2024Wav2Vec}. While end-to-end fine-tuning often achieves strong within-corpus performance, the resulting models can exploit arbitrary label-correlated cues, making it difficult to determine whether performance reflects pathology-relevant information or adaptation to dataset-specific characteristics. In contrast, frozen representations with lightweight classifiers allow a direct assessment of what information is already encoded in pretrained speech models. Because current SSL models are trained almost exclusively on healthy speech, it remains unclear whether they capture PD-related speech characteristics without task-specific adaptation.

This question is particularly important because SSL speech representations encode many factors unrelated to disease, including corpus identity \cite{bioengineering10111316}, speaker identity, and gender \cite{chiu2026largescaleprobinganalysisspeakerspecific}. Existing work has primarily sought to suppress these factors through techniques such as domain-adversarial training \cite{siniukov2025generalizablespeechmarkerparkinsons}. Rather than attempting to remove presumed confounds, we ask a different question: \textbf{Under clinically relevant distribution shifts, is the discriminative signal learned from frozen SSL representations specific to PD?} Prior work typically evaluates PD speech classifiers only against healthy controls \cite{yokoi2023analysis, dhanalakshmi2024speech, Quatra2025BilingualDD, Hossain2023MachineLC, Alhawiti2025MultiModalDH} leaving it unclear whether they capture PD-specific patterns or more general markers of neurodegeneration.

To answer this question, we introduce a five-scenario evaluation framework that begins with a no-shift reference condition (REF) using intra-corpus cross-validation, and progressively introduces increasingly challenging distribution shifts. Starting from REF, we evaluate take–re-take recordings within the same corpus (S1), recording-condition shifts (S2), language shifts (S3), task shifts while holding language constant (S4), and combined language-and-task shifts (S5). The probing layer is selected separately for each corpus, backbone, and task in the REF setting and then fixed throughout all transfer experiments. Across all conditions, we use frozen SSL speech encoders with a logistic regression probe to measure the linearly decodable information in the representations. We evaluate three PD speech corpora in Spanish (ES), German (DE), and Czech (CZ). We then transfer PD-trained classifiers to the independent, cognitively characterised German TREND cohort to compare predictions for participants with PD and dementia.

\section{\textbf{Methods}}

\subsection{Datasets}

\subsubsection{Special Session Corpora}

We used three cross-sectional corpora from the session organisers \cite{hernandez2026language}. Each corpus contains the same three tasks, recorded in one session per participant: i) DDK, ii) READ and iii) VOWEL. Characteristics of all three cohorts are summarised in Table \ref{tab:pub_corpus}.
\begin{table}[t]
\small % Reduces font size for better fit
\setlength{\tabcolsep}{2.5pt} % Further reduced from 4pt
\renewcommand{\arraystretch}{0.9} % Tighten rows
\caption{Cohort overview across special session corpora. PD: Parkinson's disease;
HC: healthy controls.}
\label{tab:pub_corpus}
\renewcommand{\arraystretch}{1.15}
\setlength{\tabcolsep}{3pt}
\footnotesize
\begin{threeparttable}
\begin{tabularx}{\columnwidth}{@{}cl *{4}{>{\centering\arraybackslash}X}@{}}
\toprule
& \textbf{Field} & \textbf{ES} & \textbf{ES-e} & \textbf{DE} & \textbf{CZ} \\
\midrule
\multicolumn{2}{@{}l}{N (PD/HC)} & 50/50 & 20/20 & 88/88 & 50/50 \\
\midrule
\multirow{4}{*}{\rotatebox[origin=c]{90}{\textbf{PD}}}
 & Sex (M/F)                       & 25/25          & 9/11            & 47/41           & 30/20 \\
 & Age                             & $61.1 \pm 9.6$ & $61.2 \pm 14.3$ & $66.5 \pm 9.0$  & $63.4 \pm 9.5$ \\             
 & MDS-UPDRS-III\tnote{$\dagger$}  & $37.6 \pm 18.2$ & $40.7 \pm 22.2$  & $22.7 \pm 10.9$ & $20.1 \pm 10.9$ \\
 & Time post diag. & $10.7 \pm 9.2$ & $9.2 \pm 6.4$    & $0.6 \pm 0.5$   & $6.7 \pm 4.7$ \\
\cmidrule(l){2-6}
\multirow{3}{*}{\rotatebox[origin=c]{90}{\textbf{HC}}}
 & Sex (M/F)  & 25/25          & 11/9            & 44/44           & 30/20 \\
 & Age        & $61.0 \pm 9.5$ & $62.6 \pm 10.0$ & $63.2 \pm 14.0$ & $61.6 \pm 11.2$ \\
\bottomrule
\end{tabularx}
\begin{tablenotes}
  \footnotesize
  \vspace{3pt}
  \item \tnote{$\dagger$} For the Spanish corpus the newer UPDRS-III version is
  used (range 0--132) versus the older version (range 0--108).
\end{tablenotes}
\end{threeparttable}
\end{table}

\paragraph{Czech (CZ)} This corpus \cite{CZ_Dataset} contains 100 participants (50 PD, 50 HC), each completing two full repetitions of all three tasks within one session. We treat the two repetitions as separate samples for Scenario~1 (S1, +Re-Take). For the VOWEL task, we discard non-/a/ recordings to avoid contamination, leaving 89 participants (48 HC, 41 PD). 
\paragraph{Spanish (ES, ES-e)} This corpus \cite{ES_Dataset} has two participant-disjoint subsets recorded under different acoustic conditions. The primary subset (ES) contains 100 participants (50 PD, 50 HC), each providing recordings for all three tasks. We apply the same VOWEL filtering as in CZ, retaining only participants with two /a/-vowel recordings; no participants were removed in this subset. The second subset (ES-e) contains 40 participants (20 PD, 20 HC) recorded in a more acoustically challenging setting, with only DDK and READ recordings; these were used for Scenario~2 (S2, +Condition).
\paragraph{German (DE)} This corpus \cite{German_Data} consists of 176 participants (88 PD and 88 HC), each providing one recording for each of the three tasks. This corpus anchors the cross-pathology transfer setting (S4, +Task), as it shares the same language as the TREND cohort.
\begin{table}[]
\small % Reduces font size for better fit
\setlength{\tabcolsep}{2.5pt} % Further reduced from 4pt
\renewcommand{\arraystretch}{0.9} % Tighten rows
\caption{TREND cohort overview across two diagnosed conditions with their matched HC groups.}
\label{tab:trend_corpus}
\renewcommand{\arraystretch}{1.2} % Slightly more spacing for readability
\setlength{\tabcolsep}{2pt} % Adjusting padding to fit 7 columns
\footnotesize
\begin{threeparttable}
\begin{tabularx}{\columnwidth}{@{} l *{6}{>{\centering\arraybackslash}X} @{}}
\toprule
\textbf{} & \multicolumn{2}{c}{\textbf{Dementia}} & \multicolumn{2}{c}{\textbf{PD}}  \\
\cmidrule(lr){2-3} \cmidrule(lr){4-5} \cmidrule(lr){6-7}
 & \textbf{Diagnosed} & \textbf{HC} & \textbf{Diagnosed} & \textbf{HC}  \\
\midrule
N               & 36            & 36            & 18            & 18                              \\
Sex (M/F)       & 23/13         & 23/13         & 16/2          & 16/2                    \\
Age             & $81.4 \pm 7.1$ & $78.2 \pm 6.2$ &  $76.9 \pm 5.5$ & $76.3 \pm 4.9$ &  \\
Age range       & [66, 92]      & [66, 89]      & [68, 87]      & [68, 84]          \\
Education (y)   & $15.2 \pm 2.9$ & $15.5 \pm 3.1$ &$15.6 \pm 2.4$ & $15.3 \pm 2.6$  \\
CERAD total score  & $62.9 \pm 13.0$ &$84.6 \pm 8.7$&$74.9 \pm 10.0$ &$84.6 \pm 8.5$\\
MMSE&$24.9 \pm 2.8$&$28.3 \pm 1.6$&$27.2 \pm 1.5$&$27.8 \pm 2.2$\\

\bottomrule
\end{tabularx}
\end{threeparttable}
\end{table}

\begin{figure}
    \centering
    \includegraphics[width=0.99\linewidth]{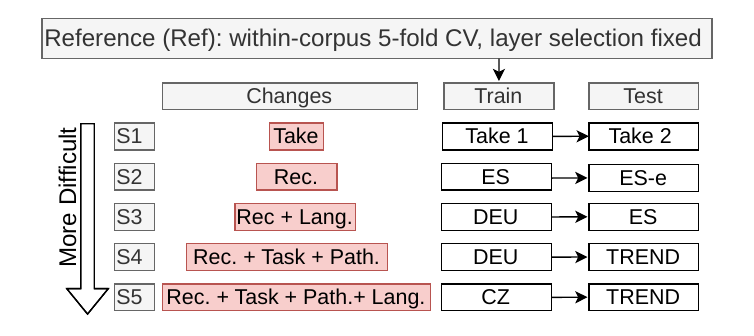}
    \caption{\textbf{Five scenarios for evaluating cross-lingual transfer in speech-based PD detection.}
REF uses 5-fold intra-corpus cross-validation to select the SSL backbone layer, which is then frozen while the binary PD classifier is trained on the full training corpus and evaluated on the target corpus without adaptation. Scenarios (S1--S5) introduce progressively larger train--test distribution shifts. The \textit{Train} and \textit{Test} columns show one representative train$\rightarrow$test example per scenario. The \textit{Changes} column summarises the factors that differ from REF. Abbreviations: Rec.\ = recording condition; Lang.\ = language; Path.\ = pathology; DE = German corpus; ES = Spanish corpus; ES-e = Spanish corpus under noisy recording conditions; CZ = Czech corpus; TREND = in-house German cohort including PD and dementia patients.}
    \label{fig:workflow}
\end{figure}

\subsubsection{In-house TREND corpus}
To extend the diagnostic scope beyond PD, we utilise the TREND \cite{trendstudie2026} dataset \footnote{Ethics approval granted by: ANONYM (approval number anonymized for the purposes of blind review)}, including recordings from non-overlapping dementia and PD participants. HCs were selected from individuals who did not convert to PD, dementia, or prodromal mild cognitive impairment (MCI) during the study period. We matched HCs to clinical subjects using a sex-matched Hungarian algorithm \cite{crouse2016implementing, virtanen2020scipy}; applying tiered tolerances for age and education. This resulted in the TREND corpus, comprising dementia (36 Dem, 36 HC) and PD (18 PD, 18 HC) groups, as summarised in Table~\ref{tab:trend_corpus}. Unlike the special-session corpora, the TREND protocol included five CERAD-NB+ tasks and one MMSE assessment. The relationship between acoustic features and cognitive performance across these tasks has previously been demonstrated by the authors~\cite{kopar2026binaryspeechrepresentationscognitive}, and we follow the same preprocessing protocol.
 \subsection{Feature Extraction}
\label{sec:features}
To enable a rigorous comparison across feature representations, we extract utterance-level representations from handcrafted acoustic features and frozen SSL backbones spanning three axes of variation: (i) capacity, (ii) ASR fine-tuning, and (iii) pretraining language.
\\ \textbf{1. Handcrafted features:} We extract the extended Geneva Minimalistic Acoustic Parameter Set (\textcolor{egemaps}{eGeMAPS})~\cite{Eyben16} using openSMILE~\cite{eyben2010opensmile}, yielding 88-dimensional utterance-level representations. The feature set includes prosodic (e.g., pitch, loudness, speech rate), spectral/articulatory (e.g., formants, MFCCs, spectral slope), and voice-quality (e.g., jitter, shimmer, harmonics-to-noise ratio) descriptors, providing theory-driven and interpretable features.\\
\textbf{2. Capacity (Base vs.\ Large):}  We use Base (\textcolor{colHuBB}{HuBERT-B}, \textcolor{colWavLMB}{WavLM-B}) and Large (\textcolor{colHuBL}{HuBERT-L}, \textcolor{colWavLML}{WavLM-L}) variants of HuBERT~\cite{hsu2021hubert} and WavLM~\cite{chen2022wavlm}.\\
\textbf{3. Fine-tuning (pretrained vs.\ ASR-fine-tuned):} We include pretrained \textcolor{colW2V2B}{W2V2-B}~\cite{baevski2020wav2vec} and \textcolor{colHuBL}{HuBERT-L} alongside their ASR-fine-tuned counterparts (\textcolor{colW2V2BFT}{W2V2-B-FT} and \textcolor{colHuBLFT}{HuBERT-L-FT}).\\
\textbf{4. Pretraining language (monolingual vs.\ multilingual):} We include the English-only pretrained model \textcolor{colW2V2B}{W2V2-B} alongside the multilingual pretrained models \textcolor{colXLSR}{XLS-R}~\cite{babu2022xlsr} (128 languages) and \textcolor{colMMS}{MMS}~\cite{pratap2023mms} (1,406 languages).\\
For each model, we extract frame-level representations from every transformer layer (including CNN output as layer 0) and obtain utterance-level embeddings by temporal mean pooling. We evaluate each corpus–task–backbone–layer combination using repeated 5-fold cross-validation with 10 random seeds. Within each fold, embeddings are standardised using the training split before fitting a logistic regression classifier~\cite{scikit-learn}. The layer with the highest mean balanced accuracy (BA) is selected in REF and used for all subsequent scenarios.

\subsection{Five Scenarios (S1--S5)}
\label{subsec:diag-lad}

To assess robustness in cross-lingual, cross-corpus PD detection, we define five scenarios that progressively introduce additional sources of distribution shift, shown in Figure \ref{fig:workflow}. All scenarios are anchored to within-corpus reference setting (REF, intra-corpus CV). For each combination, the same layer selected under REF is used across all scenarios. This ensures that performance differences reflect distribution shift rather than changes in layer selection or representation choice.

\begin{itemize}
    \item \textbf{S1 (+Re-Take)} isolates the mildest distribution shift. It reuses the same speaker-disjoint folds as REF, replacing only the test recordings with a second take while keeping the corpus, task, language, and recording condition fixed. The binary PD classification performance change $(BA_{\text{S1}} - BA_{\text{REF}})$ therefore reflects only repeated acquisition from the same speakers within a session.

    \item \textbf{S2 (+Condition)} adds recording-condition shift. We use the Spanish ES (clean) and ES-e (noisy) corpora, which contain non-overlapping participants performing the same READ and DDK speech tasks. Robustness is evaluated through bidirectional cross-condition transfer (train on ES, test on ES-e, and vice versa), quantifying sensitivity to recording conditions.

    \item \textbf{S3 (+Language)} introduces cross-lingual, cross-corpus transfer. Models are trained on a single source corpus and evaluated independently on two target corpora (e.g., $DE \rightarrow ES$ and $DE \rightarrow CZ$). This scenario quantifies generalisation across corpora.

    \item \textbf{S4 (+Task)} holds language fixed while introducing task shift. Models are transferred by inference only to the TREND corpus, which contains unseen clinical protocols (CERAD NB+ and MMSE). Beyond robustness to a new speech task, this scenario tests whether classifier outputs capture PD-specific signal.

    \item \textbf{S5 (+Task +Language)} combines the shifts introduced in S3 and S4. Models are trained on ES or CZ and transferred to the TREND corpus, introducing both language and task shift simultaneously. This represents the most challenging, deployment-oriented evaluation setting.
\end{itemize}

%==================================================
\section{\textbf{Results}}
%==================================================
\label{sec:results}
\subsection{Layer Selection}
\begin{table}[]
\small
\setlength{\tabcolsep}{2.5pt}
\renewcommand{\arraystretch}{0.9}
\caption{\textbf{Argmax layer per backbone.} Optimal layer (argmax of mean balanced accuracy across layers; 10 seeds $\times$ 5-fold CV) per backbone, task, and corpus. $\sigma$ is the standard deviation of \emph{normalised} layer depth (layer $\div$ total number of layers) across the three corpora (DE, ES, CZ) at fixed backbone and task.}
\label{tab:std-layer}
\footnotesize
\begin{threeparttable}
\begin{tabularx}{\columnwidth}{@{}l | *{3}{>{\centering\arraybackslash}X} c | *{3}{>{\centering\arraybackslash}X} c | *{3}{>{\centering\arraybackslash}X} c@{}}
\toprule
 & \multicolumn{4}{c|}{\textbf{DDK}} & \multicolumn{4}{c|}{\textbf{READ}} & \multicolumn{4}{c}{\textbf{VOWEL}} \\
 & DE & ES & CZ & $\sigma$ & DE & ES & CZ & $\sigma$ & DE & ES & CZ & $\sigma$ \\
\midrule
W2V2-B   & 5  & 0  & 5  & 0.20          & 4  & 6  & 5  & 0.07          & 0  & 1  & 3  & 0.10 \\
W2V2-FT  & 7  & 0  & 5  & \gmax{0.25}   & 6  & 4  & 8  & \gmax{0.14}   & 0  & 1  & 9  & \gmax{0.34} \\
\midrule
HuB-B    & 9  & 0  & 7  & 0.32          & 6  & 10 & 7  & 0.14          & 3  & 4  & 4  & 0.04 \\
HuB-L    & 24 & 1  & 15 & \gmax{0.39}   & 24 & 3  & 18 & \rmax{0.37}   & 3  & 2  & 21 & \rmax{0.36} \\
HuB-L-FT & 15 & 1  & 11 & 0.25          & 18 & 8  & 13 & 0.17          & 0  & 17 & 7  & 0.29 \\
\midrule
WavLM-B  & 7  & 0  & 7  & 0.27          & 9  & 5  & 6  & 0.14          & 5  & 5  & 2  & 0.12 \\
WavLM-L  & 24 & 1  & 22 & \rmax{0.43}   & 21 & 2  & 9  & \gmax{0.33}   & 13 & 9  & 16 & \gmax{0.12} \\
\midrule
XLS-R    & 16 & 10 & 16 & \gmax{0.12}   & 20 & 3  & 10 & 0.29          & 22 & 8  & 5  & \gmax{0.31} \\
MMS      & 9  & 13 & 10 & 0.07          & 23 & 11 & 6  & \gmax{0.30}   & 14 & 20 & 12 & 0.14 \\

\bottomrule
\end{tabularx}
\begin{tablenotes}
  \footnotesize
  \vspace{3pt}
  \item Highlighting: \gmaxtext{group max} (per backbone family) and \rmaxtext{column max}.
\end{tablenotes}
\end{threeparttable}
\end{table}
\begin{figure*}
    \centering
    \includegraphics[width=0.99\linewidth]{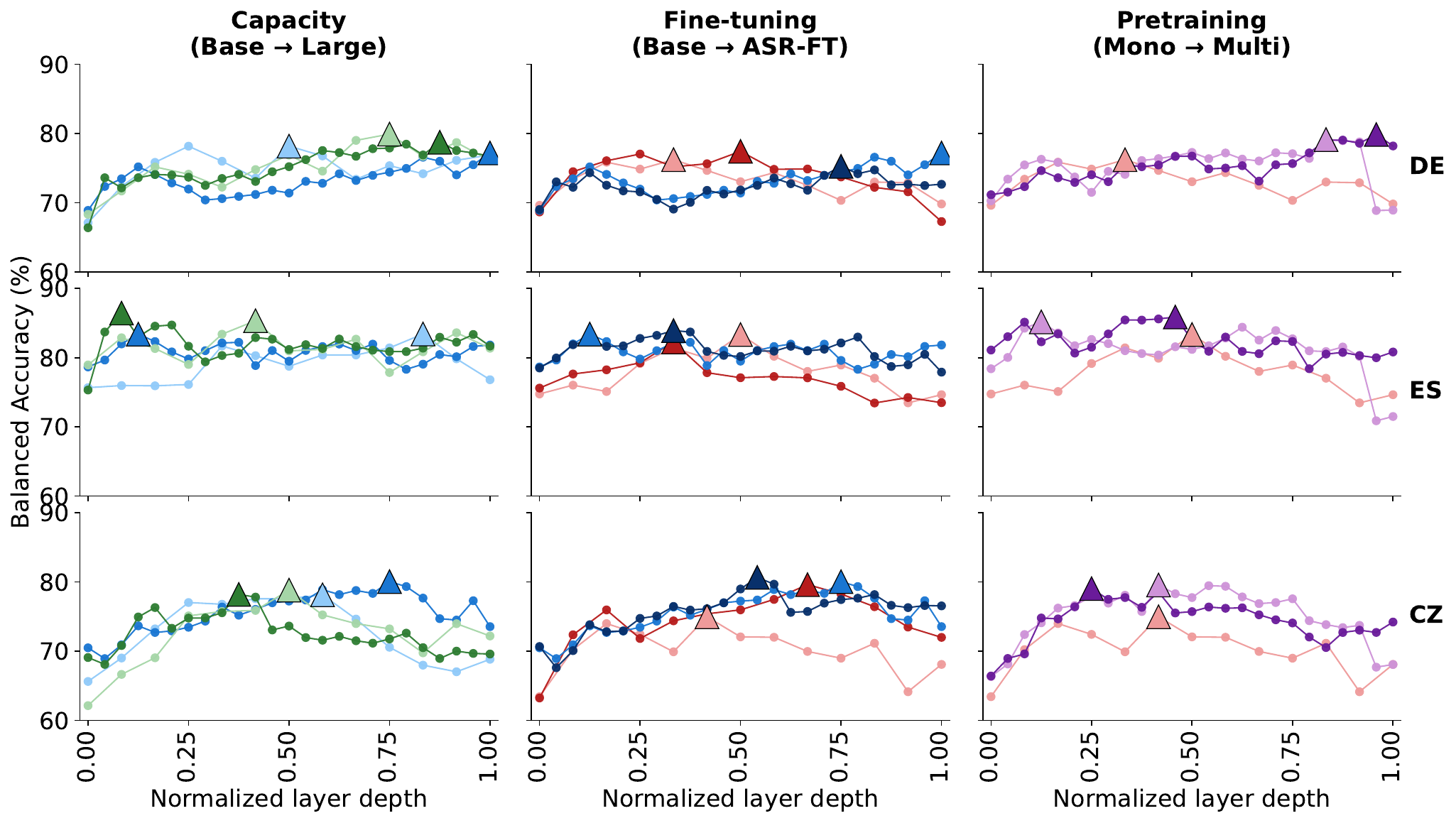}
    \caption{\textbf{Layer selection results for the READ task in the REF setting, where the selected backbone layer is fixed for all downstream transfer experiments.} The three plots compare: i) \textbf{Capacity}: \textcolor{colHuBB}{HuBERT-B} vs.\ \textcolor{colHuBL}{HuBERT-L} and \textcolor{colWavLMB}{WavLM-B} vs.\ \textcolor{colWavLML}{WavLM-L}; ii) \textbf{Fine-tuning}: \textcolor{colW2V2B}{W2V2-B} vs.\ \textcolor{colW2V2BFT}{W2V2-B-FT} and \textcolor{colHuBL}{HuBERT-L} vs.\ \textcolor{colHuBLFT}{HuBERT-L-FT}; iii) \textbf{Pretraining}: \textcolor{colW2V2B}{W2V2-B} vs.\ \textcolor{colXLSR}{XLS-R} and \textcolor{colMMS}{MMS}. The READ task is shown because it achieved the highest overall balanced accuracy (BA).}
    \label{fig:layer-selection}
\end{figure*}
\textbf{Optimal layer selection is corpus-dependent and least stable for large backbones.}
Table~\ref{tab:std-layer} reports each selected layer across 10 seeds and 5-fold cross-validation, together with the standard deviation ($\sigma$) of normalised optimal-layer depth. Optimal layer choice varies substantially across corpora, especially for larger backbones. On DDK, WavLM-L selects layer 24 for DE, layer 1 for ES, and layer 22 for CZ ($\sigma = 0.43$), with a similar pattern across tasks (e.g., HuBERT-L reaches $\sigma = 0.37$ on READ and $\sigma = 0.36$ on VOWEL). 

Figure~\ref{fig:layer-selection} shows BA across layers for READ. Layer-wise performance appears to vary by corpus rather than following a uniform trend. For DE, BA tends to increase with depth before plateauing. For ES, it remains relatively stable across layers. For CZ, performance appears to follow a weak unimodal pattern, with mid-depth layers performing slightly better than the deepest ones. These trends are broadly consistent across backbone variance axis (capacity, fine-tuning and pretraining), suggesting that layer-wise behaviour is driven more by corpus than by architecture.
%%%%%%%%%%%%%%%%%%%%%%%%%%%%%%%%%%%%%%%%%%%%%%%%%%%%%%%
\subsection{Scenario 1 (+Re-Take)} 
\begin{figure}
    \centering
    \includegraphics[width=0.99\linewidth]{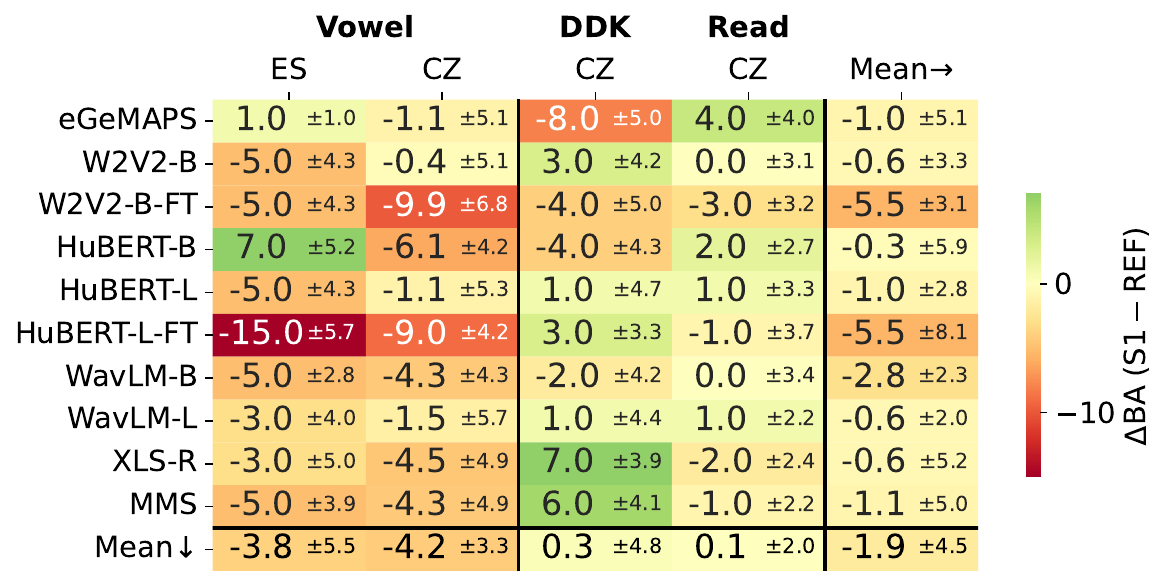}
\caption{\textbf{S1 (+Re-Take) Results:} Change in balanced accuracy relative to REF ($\Delta\mathrm{BA}=\mathrm{BA}_{\mathrm{S1}}-\mathrm{BA}_{\mathrm{REF}}$). S1 differs from REF only in the test recordings, which are replaced by each participant's second recording; training data and participant splits are unchanged. Values close to zero indicate robust performance across recording takes. Results are reported as mean $\pm$ bootstrapped SD ($n=2000$ test-set resamples). Columns correspond to speech task and corpus. The rightmost column (\textit{Mean}) reports the average $\Delta$BA for each backbone across all tasks and corpora, while the bottom row reports the average $\Delta$BA for each corpus--task combination across backbones. Cell colour encodes the sign and magnitude of $\Delta$BA.}
    \label{fig:s1-results}
\end{figure}

\textbf{Performance across recording takes depends on the interplay between speech task and backbone.} Figure~\ref{fig:s1-results} reports the change in binary PD classification balanced accuracy relative to REF $\Delta \mathrm{BA} = \mathrm{BA}_{\mathrm{S1}} - \mathrm{BA}_{\mathrm{REF}}$, where only the test recordings are replaced by each participant's second take. Thus, values close to zero indicate robustness to repeated recordings.

Overall, mean performance change is small ($-1.9 \pm 4.5$ BA points, bottom right). READ shows the smallest mean performance change ($0.1 \pm 2.0$), followed by DDK ($0.3 \pm 4.8$). DDK is also the only task where both multilingual backbones improve consistently, with XLS-R gaining $7.0 \pm 3.9$ BA points and MMS $6.0 \pm 4.1$.
VOWEL exhibits the greatest sensitivity to recording re-takes. Across all models, the average performance decreases by $3.8 \pm 5.5$ BA points on ES and $4.2 \pm 3.3$ BA points on CZ. However, the effect varies considerably across SSL backbones and corpora. For instance, HuBERT-B improves on ES VOWEL by $7.0 \pm 5.2$ BA points, while its performance drops by $6.1 \pm 4.2$ BA points on CZ. The largest degradations are observed for the ASR-fine-tuned models: HuBERT-L-FT decreases by $15.0$ BA points on ES, and W2V2-B-FT decreases by $9.9$ BA points on CZ. Overall, these results show that the impact of recording re-takes is strongly dependent on both the backbone and the corpus, with VOWEL exhibiting the highest variability.
\begin{figure}
    \centering
    \includegraphics[width=0.99\linewidth]{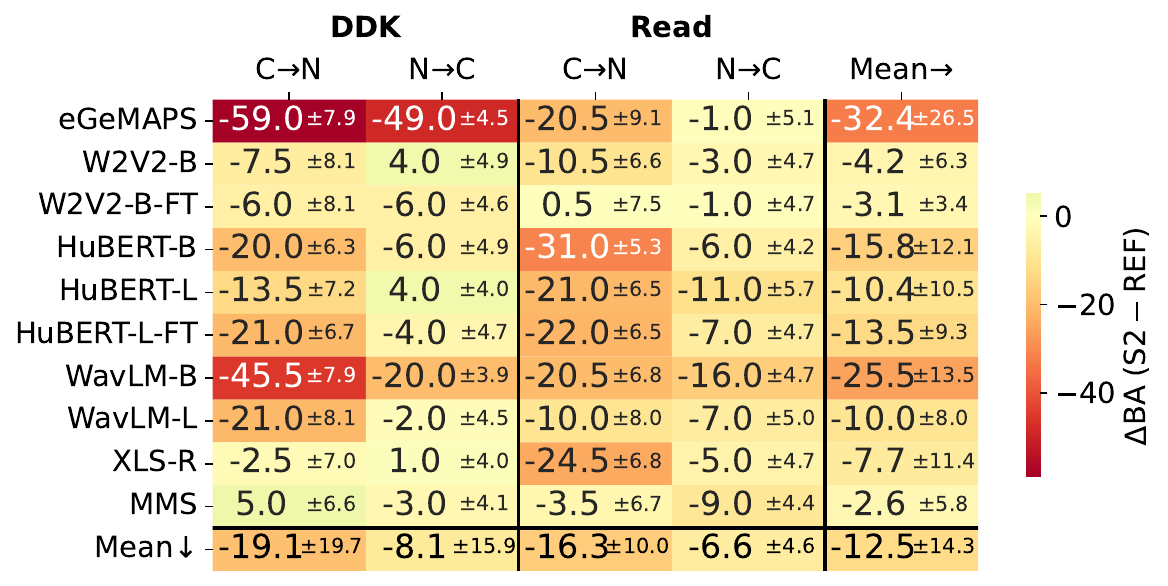}
    \caption{\textbf{S2 (+Condition) Results:} Change in balanced accuracy relative to REF ($\Delta\mathrm{BA}=\mathrm{BA}_{\mathrm{S2}}-\mathrm{BA}_{\mathrm{REF}}$) under recording-condition shift between clean (C) ES and noisy (N) ES-e corpora with non-overlapping speakers. C$\rightarrow$N denotes training on clean and testing on noisy data; N$\rightarrow$C denotes the reverse. Results are reported as mean $\pm$ bootstrapped SD ($n=2000$ resamples). Columns correspond to speech task and corpus. The rightmost column (\textit{Mean}) averages $\Delta$BA per backbone across all tasks and corpora, while the bottom row averages across backbones per corpus--task pair. Cell colour encodes the sign and magnitude of $\Delta$BA.}
    \label{fig:s2-results}
\end{figure}

\begin{figure}[b]
    \centering
    \includegraphics[width=0.99\linewidth]{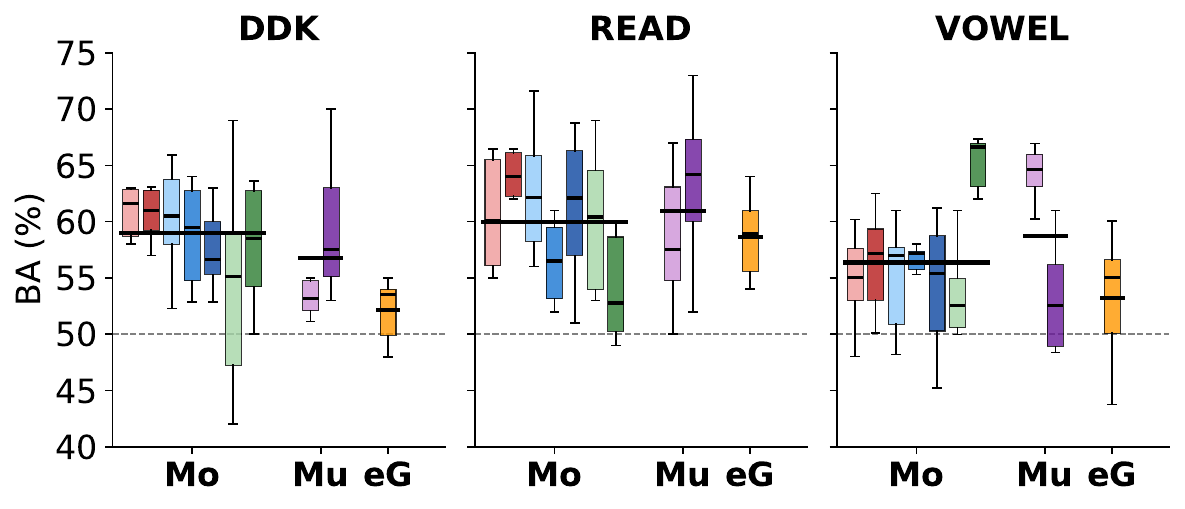}
    \caption{\textbf{S3 (+Language) Results:} Seven monolingual pretrained SSL backbones (\textcolor{colW2V2B}{W2V2-B}, \textcolor{colW2V2BFT}{W2V2-B-FT}, \textcolor{colHuBB}{HuBERT-B}, \textcolor{colHuBL}{HuBERT-L}, \textcolor{colHuBLFT}{HuBERT-L-FT}, \textcolor{colWavLMB}{WavLM-B}, \textcolor{colWavLML}{WavLM-L}), two multilingual SSL backbones (\textcolor{colXLSR}{XLS-R}, \textcolor{colMMS}{MMS}), and handcrafted \textcolor{egemaps}{eGeMAPS} features. Horizontal lines indicate group-wise mean values.}
    \label{fig:s3-results}
\end{figure}

\begin{figure*}
    \centering
    \includegraphics[width=0.99\linewidth]{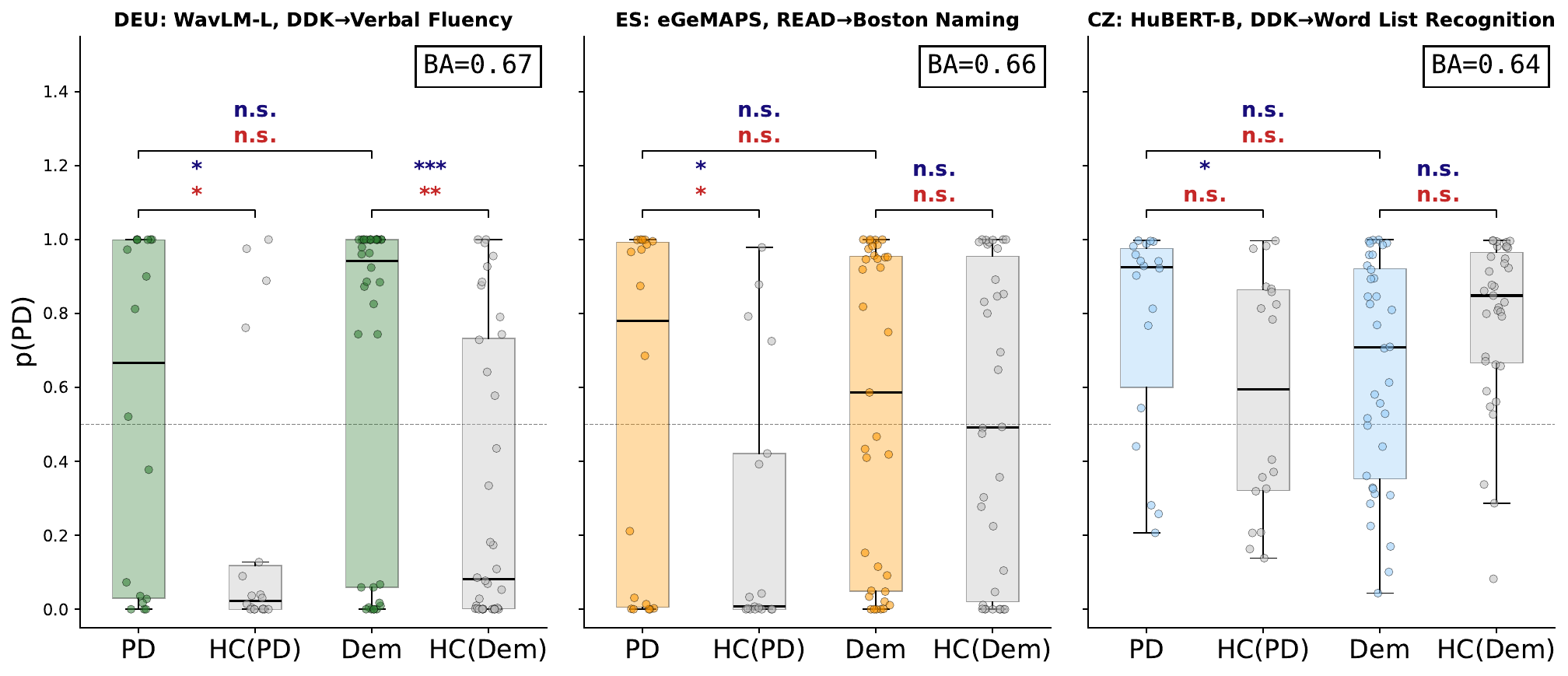}
\caption{\textbf{S4 (+Task) and S5 (+Task + Language) Results:} Predicted PD probability distributions for the highest-balanced-accuracy (BA) transfer combination of each training corpus. Layers were selected using only within-corpus REF data; no TREND data was used during layer selection. Each panel shows $p(\mathrm{PD})$ across the four TREND cohort groups. Box colours correspond to the respective SSL backbone. Significance brackets indicate \textcolor{mwu}{Mann--Whitney U} and \textcolor{perm}{permutation test} results ($n=1000$); significance levels are denoted by $^{***}p<0.001$, $^{**}p<0.01$, and $^{*}p<0.05$. Abbreviations: PD = Parkinson's disease; HC(PD) = matched healthy control; Dem = dementia; HC(Dem) = matched healthy control.}
    \label{fig:trend-s4}
\end{figure*}
%%%%%%%%%%%%%%%%%%%%%%%%%%%%%%%%%%%%%%%%%%%%%%%%%%%%%%%%%%%%%%%%%
\subsection{Scenario 2 (+Condition)}
\textbf{Cross-condition transfer is asymmetric: training on noisy recordings transfers better to clean recordings than vice versa.} Building on S1 (+Re-Take), S2 (+Condition) introduces a recording-condition shift by training and testing across the clean (ES) and noisy (ES-e) variants of the same Spanish corpus with non-overlapping speakers, while keeping language and task fixed. Figure~\ref{fig:s2-results} shows $\Delta$BA (S2 $-$ REF) with bootstrapped standard deviations for both transfer directions (Clean$\rightarrow$Noisy and Noisy$\rightarrow$Clean). Unlike the modest overall degradation in S1 ($-1.9 \pm 4.5$), recording-condition mismatch causes a substantially larger performance drop, with an overall mean of $-12.5 \pm 14.3$ (bottom right).

Two patterns emerge. First, transfer is consistently asymmetric: training on noisy data generalises better to clean recordings than the reverse. Across backbones, mean $\Delta$BA improves from $-19.1 \pm 19.7$ to $-8.1 \pm 15.9$ on DDK and from $-16.3 \pm 10.0$ to $-6.6 \pm 4.6$ on READ. The largest improvements are observed for base models, WavLM-B on DDK ($-45.5 \rightarrow -20.0$) and HuBERT-B on READ ($-31.0 \rightarrow -6.0$). MMS and W2V2-B-FT are the most condition-robust overall (row means $-2.6 \pm 5.8$ and $-3.1 \pm 3.4$, respectively).

Second, the handcrafted eGeMAPS baseline is the most condition-sensitive ($-32.4 \pm 26.5$), with the largest individual degradations ($-59.0$ and $-49.0$ on DDK), highlighting the greater overall mean robustness of SSL representations to recording-condition mismatch.

\subsection{Scenario 3 (+Language)} \textbf{Backbone performance depends on the recorded task.}
S3 (+Language) introduces a cross-lingual evaluation setting in which each backbone is trained on one corpus and tested individually on the other two, yielding six balanced accuracy (BA) values per backbone and task. Figure~\ref{fig:s3-results} reports binary PD classification performance for three groups: monolingual (Mo) SSL models, multilingual (Mu) SSL models, and handcrafted eGeMAPS features (eG). The black horizontal lines indicate the mean BA within each group.

Overall, eG shows lower mean performance across tasks, suggesting weaker cross-corpus transfer. Although this setup directly evaluates cross-lingual generalisation, no consistent advantage of multilingual (Mu) over monolingual (Mo) SSL backbones is observed. Consistent with earlier findings, the recording task has a pronounced effect across SSL backbones: WavLM-L, in particular, exhibits non-overlapping performance values for VOWEL versus READ. A similar pattern observed for XLS-R when comparing its VOWEL and DDK performance. Taken together, these results indicate that task and backbone choice have a stronger influence on performance than the monolingual versus multilingual distinction alone.

Across scenarios, performance relative to REF shows a progressive decline, with mean $\Delta$BA of $-1.9 \pm 4.5$ (S1-REF), $-12.5 \pm 14.3$ (S2-REF), and $-16.3 \pm 10.6$ (S3-REF), indicating increasing degradation under more challenging conditions.

\subsection{Scenarios 4 (+Task) \& 5 (+Task + Language)}
\textbf{Cross-task transfer to TREND detects disorder, not PD specifically.}
Out of 540 transfer combinations (10 backbones × 6 TREND tasks × 3 PD tasks × 3 training corpora), only 9 (6 trained on DE, 2 on ES, and 1 on CZ) achieved at least moderate separation between PD and HC, defined as $AUC > 0.6$ and $BA > 0.6$.
We therefore restrict the specificity analysis to these 9 cases, since evaluating PD-specificity is only meaningful when there is some evidence of baseline PD classification performance. Importantly, the non-specificity pattern described below is consistent across all 9 remaining combinations.

Figure~\ref{fig:trend-s4} shows the predicted PD probability distributions across the four TREND subgroups for the highest BA classifier trained on each corpus (DE, ES, CZ). BA is computed for the PD and HC(PD) comparison. All three classifiers achieve only modest separation between PD and matched controls (DE = 0.67, ES = 0.66, CZ = 0.64). However, performance does not generalise to specificity in the PD vs. dementia comparison. This is most evident in the German transfer setting (S4 +Task), where the classifier significantly separates PD from HC(PD) and dementia from HC(Dem) ($p < 0.01$), but fails to distinguish PD from dementia according to both permutation testing \cite{churchill2008naive} and the Mann–Whitney U test \cite{mann1947test}. Overall, these results suggest a disorder-general signal rather than PD-specific discrimination.

Additionally, we trained a binary logistic regression classifier from scratch to distinguish the PD from dementia, only using the demographics features as input. Age, sex, and education alone already discriminate PD from dementia with AUC values of 0.73–0.75. We than adjusted the PD probability scores for these covariates using leave-one-subject-out (LOSO) regression. The residualised performance drops to chance level (AUC = 0.50–0.55), and all confidence intervals include 0.5, supporting previous claim of non-specificity.
\section{\textbf{Discussion and Limitations}}
\label{sec:discussion}
In this study, we show that speech-based PD detection is driven more by corpus-specific factors than by PD-related motor or cognitive pathology. Under cross-lingual and cross-task transfer, models largely preserve a generic patient--healthy separation rather than disease-specific structure. This is further supported by the fact that the selected backbone layer varies substantially across corpora, especially for larger backbones.

We observe a consistent degradation in binary PD classification under increasingly severe distribution shifts. Identity-preserving re-takes. S1 (+Re-Take) have a small degradation of performance ($-1.9$ BA), while recording-condition (S2) and cross-corpus (S3) transfers lead to a larger drops ($-12.5$ and $-16.3$ BA). Cross-lingual shifts are challenging, even multilingual pretraining not showing a clear advantage. If the classifier captured robust motor impairment, greater stability across re-takes and partial cross-lingual robustness would be expected. Instead, the sensitivity across settings suggests that learned representations are not purely motoric. These findings align with prior work showing that language and corpus factors confound pathology classification \cite{hernandez2026language} and that SSL embeddings tend to cluster by dataset rather than disease \cite{ibarra2023domain}.

A key clinical implication emerges in transfer to the TREND dataset. Although transferred models moderately separate PD from healthy controls, they fail to distinguish PD from dementia. This is particularly concerning because most speech-based PD studies evaluate performance only against healthy controls, leaving diagnostic specificity to other neurological conditions largely untested. To our knowledge, this is the first systematic evaluation of cross-lingual PD classifiers in a dementia setting. While our results suggest that strong intra-corpus performance may not generalise under clinically meaningful shifts, conclusions are limited by our modest cohort sizes. We therefore interpret our findings as \emph{absence of evidence for PD-specificity}, not evidence of its absence. 

Overall, we view this work as a proof of concept for cross-disease transfer evaluation. It addresses the central question posed in our title, showing that what survives across transfer settings is primarily driven by the corpus, reflecting a general divergence from healthy speech, rather than a robust speech signature specific to Parkinson’s disease.

\newpage
\section*{Acknowledgment}

%\newpage
\bibliographystyle{IEEEtran}
\bibliography{refs}
\end{document}